\pdfoutput=1
\documentclass[11pt]{article}

\usepackage[preprint]{acl}

\usepackage{times}
\usepackage{latexsym}
\usepackage{stfloats}
\usepackage[T1]{fontenc}
\usepackage[utf8]{inputenc}

\usepackage{microtype}

\usepackage{inconsolata}

\usepackage{graphicx}

\usepackage{colortbl}
\usepackage{booktabs}
\usepackage[subtle]{savetrees}
\usepackage[textsize=tiny]{todonotes}
\usepackage{subcaption}
\usepackage{xurl}
\usepackage{fontawesome5}
\definecolor{msftGreen}{RGB}{127,186,0}
\definecolor{msftYello}{RGB}{255,185,0}
\definecolor{mypurple}{RGB}{138,43,226} 
\definecolor{msftBlack}{RGB}{0,0,0}
\definecolor{verylightpurple}{RGB}{235, 215, 255}

\definecolor{codecolor}{RGB}{70,70,180} 
\definecolor{registercolor}{RGB}{160,40,40} 
\definecolor{immediatecolor}{RGB}{0,110,80}  
\definecolor{memorycolor}{RGB}{100,50,140}
\definecolor{labelcolor}{RGB}{180,90,40}
\newcommand{\code}[1]{{\color{codecolor}\textbf{\texttt{#1}}}}
\newcommand{\reg}[1]{{\color{registercolor}\textbf{\texttt{#1}}}}
\newcommand{\imm}[1]{{\color{immediatecolor}\textbf{\texttt{#1}}}}
\newcommand{\mem}[1]{{\color{memorycolor}\textbf{\texttt{#1}}}}

\newcommand{\GG}{\textbf{\texttt{\textcolor{mypurple}{GG}}}~}
\newcommand{\GGn}{\textbf{\texttt{\textcolor{mypurple}{GG}}}}

\title{
    \raisebox{-0.5cm}{\includegraphics[width=1.2cm]{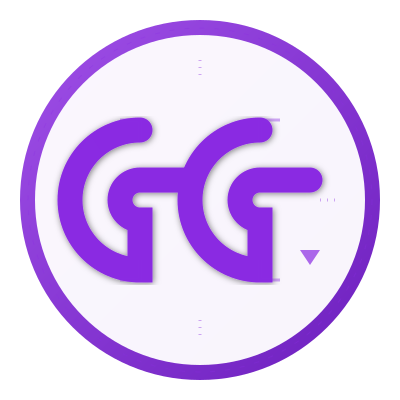}}  % Adjust vertical position
    \hspace{-0.05cm} \textcolor{mypurple}{G}uaranteed \textcolor{mypurple}{G}uess: A Language Modeling Approach for \\
    \hspace{1.0cm} CISC-to-RISC Transpilation with Testing Guarantees
}

\author{ 
 Ahmed Heakl$^{\spadesuit}$,
 Sarim Hashmi$^{\spadesuit}$,
 Chaimaa Abi$^{\spadesuit}$ \\
 \textbf{Celine Lee$^{\heartsuit}$, 
 Abdulrahman Mahmoud}$^{\spadesuit}$ \\
 $^{\spadesuit}$MBZUAI \quad $^{\heartsuit}$Cornell University \\
 \texttt{\{ahmed.heakl, sarim.hashmi, abdulrahman.mahmoud\}@mbzuai.ac.ae} \\
 \url{https://ahmedheakl.github.io/Guaranteed-Guess/}
}

\begin{document}
\maketitle

\begin{abstract}
The hardware ecosystem is rapidly evolving, with increasing interest in translating low-level programs across different \textit{instruction set architectures} (ISAs) in a quick, flexible, and correct way to enhance the portability and longevity of existing code. A particularly challenging class of this transpilation~\footnote{\textit{We use ``transpilation''} to describe the task of translating code between assembly languages.} problem is translating between complex- (CISC) and reduced- (RISC) hardware architectures, due to fundamental differences in instruction complexity, memory models, and execution paradigms. In this work, we introduce \GG (\textcolor{mypurple}{G}uaranteed \textcolor{mypurple}{G}uess), an ISA-centric transpilation pipeline that combines the translation power of pre-trained large language models (LLMs) with the rigor of established software testing constructs. Our method generates candidate translations using an LLM from one ISA to another, and embeds such translations within a software-testing framework to build quantifiable confidence in the translation. We evaluate our \GG approach over two diverse datasets, enforce high code coverage ($>$98\%) across unit tests, and achieve functional/semantic correctness of 99\% on HumanEval programs and 49\% on BringupBench programs, 
respectively. Further, we compare our approach to the state-of-the-art Rosetta 2 framework on Apple Silicon, showcasing 1.73$\times$ faster runtime performance, 1.47$\times$ better energy efficiency, and 2.41$\times$ better memory usage for our transpiled code, demonstrating the effectiveness of \GG for real-world CISC-to-RISC translation tasks. We will open-source our codes, data, models, and benchmarks to establish a common foundation for ISA-level code translation research.
\end{abstract}

\section{Introduction}
The modern hardware landscape is undergoing a fundamental transformation. As Moore’s Law slows and Dennard scaling ends~\cite{dennard_scaling, moores_law}, the demand for energy-efficient, high-performance architectures has accelerated, particularly with the rise of machine learning (ML) applications~\cite{horowitz20141, jouppi2017datacenter}. Hyperscalers are increasingly constrained by power and thermal limits~\cite{patterson2021carbon, gupta2021chasing}, prompting a reevaluation of datacenter infrastructure.

A major outcome of this shift is the growing adoption of ARM-based processors. Historically dominant in mobile and edge devices due to their RISC-based, low-power design, ARM CPUs were largely absent from datacenters because of their performance gap with x86 (a CISC architecture)~\cite{blem2013power}. However, this gap has narrowed significantly: ARM-based chips now match x86 on many benchmarks~\cite{cloudpanel_arm_servers} and deliver superior energy efficiency~\cite{ionos2024arm}. In 2024, x86 designs dominated over 80\% of data center servers~\cite{arm2025reuters}, but ARM predicts that its share will reach 50\% by the end of 2025~\cite{maruccia2025arm}. Industry adoption supports this trend, with ARM-based systems like NVIDIA’s Grace CPU~\cite{nvidia2024grace}, Amazon’s Graviton~\cite{nextplatform_graviton3}, and Microsoft’s ARM-compatible OS stack~\cite{verma2024exploring} accelerating deployment.

This rapid hardware transition introduces a significant software gap. Legacy binaries compiled for x86 often lack source code and cannot be recompiled for ARM. While solutions like Apple's Rosetta 2~\cite{rosetta2} and QEMU's emulation service~\cite{qemu} provide runtime virtualization, they introduce memory and performance overheads. Compilers struggle to retarget opaque binaries~\cite{he2018debin}, and decompilation-based approaches are fragile or legally restricted~\cite{wang2024evaluating}. A scalable, accurate, and architecture-aware binary-to-binary translation solution remains elusive.

In this work, we introduce \textit{Guaranteed Guess} (\GGn), an assembly-to-assembly transpiler that translates x86 binaries (CISC) into efficient ARM or RISC-V (RISC) equivalents using a custom-trained large language model (LLM). Our approach is \textit{open-source}, avoids the \textit{virtualization tax} by generating native ARM/RISC-V assembly, and directly supports legacy binaries \textit{without decompilation}.

Transpiling across ISAs is non-trivial. CISC and RISC architectures differ in register-memory semantics, instruction complexity, and binary length, x86 instructions are fewer but more expressive, while RISC requires longer, register-centric code sequences. These differences must be learned implicitly by the model, which we achieve by incorporating hardware-informed design, tokenizer extensions, and context-aware training.

Our approach builds high-accuracy LLM-based transpilers by incorporating hardware-aware insights into the training process, enabling the model to better capture the CISC-specific patterns of x86 and generate semantically valid RISC targets such as ARM. However, unlike high-level language tasks, conventional NLP correctness proxies (e.g., BLEU, perplexity) fall short for binary translation where functional correctness is paramount. Therefore, we embed our predictions within rigorous software testing infrastructure to provide test-driven guarantees of correctness. Holistically, our paper makes the following key contributions:

\begin{enumerate}
    \item The first CISC-to-RISC transpiler, coined \GGn, built via a custom-trained, architecture-aware LM achieving a test accuracy of  99.39\% on ARMv8 and  89.93\% on RISC-V64.  
    \item A methodology to measure and build confidence into transpilation output via software testing approaches ("guaranteeing" the guess) (\S\ref{sec:gg}), including detailed analysis of correctness, errors, and hallucinations (\S\ref{sec:eval})
    \item An in-depth analysis into the inner workings of our transpiler, including hardware-informed design decisions to best train an accurate LLM model for assembly transpilation (\S\ref{sec:gg}, \S\ref{sec:results}). 
    \item We perform a case-study using our transpiler in a real-world setting, by comparing it directly to Apple Rosetta's x86 to ARM virtualization engine. Results show that \GGn's generated assembly achieves 1.73x runtime speedup while delivering 1.47x better energy efficiency and 2.41x memory efficiency (\S\ref{sec:results}).
\end{enumerate}

% \noindent The remainder of the paper is structured as follows: we first describe related work and background in assembly-level translation and LLMs (\S\ref{sec:related}), then present our data generation and training methodology (\S\ref{sec:gg}). We detail experimental setup and evaluation metrics (\S\ref{sec:eval}), analyze results across benchmarks and settings (\S\ref{sec:results}), and conclude with insights and future directions (\S\ref{sec:conc}). 

\section{Background and Related Work}\label{sec:related}

\paragraph{Virtualization and Emulation}
Emulation and assembly-level virtualization enable the execution of one ISA’s binary on a host machine for which it was not originally compiled. QEMU~\citep{qemu}, an open-source emulator, uses dynamic binary translation~\citep{sites1993binary} to translate machine code on-the-fly, offering flexibility but with performance overhead. Supported emulation currently includes x86 to ARM, amongst other ISAs. Rosetta 2~\citep{rosetta2}, Apple’s virtualization layer for macOS, combines ahead-of-time (AOT) and just-in-time (JIT) translation, providing better performance within the Apple ecosystem.

These approaches face challenges in achieving native-level performance and ensuring broad compatibility, due to the dynamic nature of execution. A transpiler approach, directly converting x86 to ARM assembly, could supplant these solutions by eliminating runtime translation overhead with a one-time translation into the host ISA. This method could address the limitations of current emulation and virtualization techniques, particularly in performance-critical scenarios, or where pre-processing is feasible, or when source code is not available (due to proprietary IP).

\paragraph{Coding with LLMs}
Language modeling approaches for code have primarily focused on understanding, generating, and translating high-level programming languages such as C++, Java, and Python~\citep{lachaux2020unsupervised,feng2020codebert,wang2021codet5,roziere2023code,liu2024deepseek}. These models demonstrate increasingly sophisticated code manipulation capabilities through self-supervised learning on vast code repositories. Models further trained with reinforcement learning have shown remarkable performance in rules-based reasoning tasks, including code~\citep{deepseekai2025deepseekr1incentivizingreasoningcapability}.
However, the resulting models struggle when applied to languages under-represented in their training sets, in particular when used to write assembly-level code, where the semantics and structure differ significantly from their high-level counterparts.

\paragraph{Neural Low-Level Programming}
Recent research demonstrates the potential of adapting LLMs to various tasks related to low-level code analysis and transformation: decompilation, binary similarity analysis, and compiler optimization.
LLM4Decompile~\citep{tan2024llm4decompile} introduced specialized language models for direct binary-to-source translation and decompiler output refinement. DeGPT~\citep{hu2024degpt} further explored decompiler enhancement through semantic-preserving transformations.
SLaDe~\citep{armengol2024slade} combines a 200M-parameter sequence-to-sequence Transformer with type inference techniques to create a hybrid decompiler capable of translating both x86 and ARM assembly code into readable and accurate C code, effectively handling various optimization levels (-O0 and -O3).
Language models have also been adapted to optimization tasks, with LLM Compiler~\citep{cummins2024meta} introducing a foundation model that supports zero-shot optimization flag prediction, bidirectional assembly-IR translation, and compiler behavior emulation. %This approach demonstrates the potential for language models to enhance compiler optimization workflows by automating complex tasks.
Binary similarity analysis has similarly benefited from language model adaptations. DiEmph~\citep{xu2023improving} addressed compiler-induced biases in transformer models, while jTrans~\citep{wang2022jtrans} incorporated control flow information into the transformer architecture. Yu et al.~\citep{yu2020order} combined BERT-based semantic analysis with graph neural networks to capture both semantic and structural properties of binary code.
While these applications have shown promising results, the use of LLMs to port efficient machine code from one machine to another, while maintaining efficiency, remains underexplored and largely unsolved. Assembly languages present unique challenges due to their under-representation in training datasets, lack of human readability, extensive length, and fundamental differences in execution models across architectures.

Guess \& Sketch~\citep{lee2024guess} introduced a neurosymbolic approach combining language models with symbolic reasoning for translating assembly code between ARMv8 and RISC-V architectures. 
In our work, we extend the neural transpiliation direction with a focus on leveraging the existing efficiency in x86 programs to transpile into efficient ARM binaries, bridging architectural differences in ISA complexity and execution models. 
Further, instead of fixing transpilations with symbolic approaches, as done in Guess \& Sketch, we focus on upfront data design and modeling methods to flexibly handle the increased scale and complexity of CISC-to-RISC transpilation.

% \newpage
\section{\textbf{\textcolor{mypurple}{G}}uaranteed \textbf{\textcolor{mypurple}{G}}uess}\label{sec:gg}

\begin{figure*}[t]
    \centering
    \includegraphics[width=0.7\linewidth]{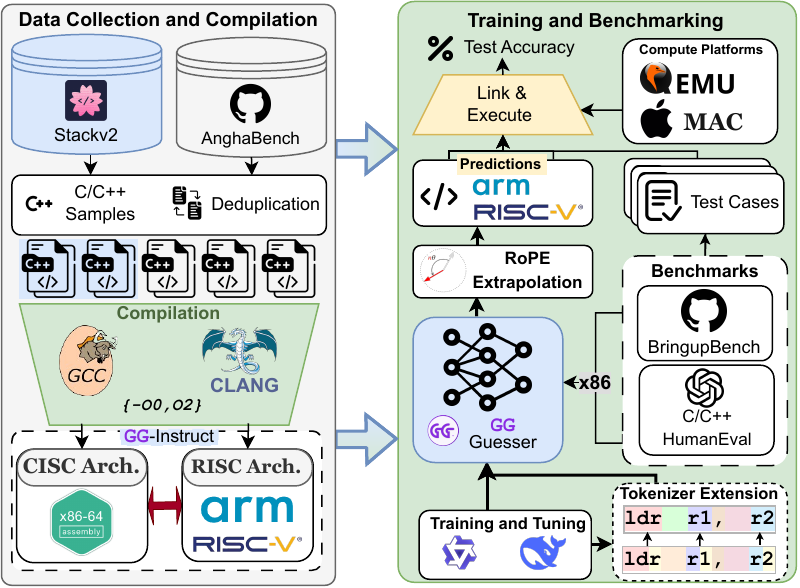}
    \caption{\textbf{\GG System Overview}. A two-stage transpilation pipeline from x86 to ARM/RISC-V. Left: Data is sourced from Stackv2 and AnghaBench, deduplicated, and compiled using both GCC and Clang to generate paired assembly (x86 $\leftrightarrow$ ARM) from C/C++. Right: A specialized LLM (\GG Guesser), trained with tokenizer extension and inferenced with RoPE extrapolation, predicts target ISA code. Predictions are evaluated via unit tests and symbolic analysis on benchmarks like HumanEval and BringupBench. The system emphasizes functional correctness, architectural alignment, and near-native performance.}
    \label{fig:transpiler-overview}
\end{figure*}

In this section, we explore the two primary components of building our \GG transpiler: data generation and model training. 

\subsection{Data Collection} 
As shown in Figure~\ref{fig:transpiler-overview}, our training dataset is derived from AnghaBench\cite{anghabench} and The Stackv2\cite{stack}. AnghaBench is a comprehensive benchmark suite that contains 1 million compilable C/C++ programs extracted from major public C/C++ repositories on GitHub. The Stack is a 3.1TB dataset of permissively licensed code in 30 languages for training and evaluating code LLMs. From these datasets, we randomly sampled 1.01M programs (16.16B tokens) from AnghaBench and 306k programs (4.85B tokens) from the stack to form our training set, equivalent to 1.32M samples. After we collected the whole samples, we removed boilerplates, deduplicated the data, and choose file that were neither too short (<10 lines) nor too long (>16k lines). These programs were then compiled for x86 (CISC) $\leftrightarrow$ ARMv8/ARMv5/RISC-V (RISC). 

Each program was compiled to both x86 (CISC) $\leftrightarrow$ ARMv8/ARMv5/RISC-V (RISC) targets under two optimization levels: \texttt{-O0} (no optimization) and \texttt{-O2} (aggressive optimization). These flags were selected to expose models to both raw, semantically transparent code (\texttt{-O0}) and real-world, performance-optimized binaries (\texttt{-O2}), enabling the model to learn both unoptimized and optimized ISA patterns. Compilation for ARMv5 and RISC-V64 was performed via cross-compilation on an Ubuntu 20.04 machine with a Ryzen 7 CPU, using \texttt{arm-linux-gnueabi-gcc}~\cite{gnueabi} and \texttt{gcc-riscv64-linux-gnu}~\cite{riscv-gcc}, respectively. ARMv8 binaries were compiled natively on an Apple M2 Pro (macOS) using \texttt{clang}~\cite{clang}, ensuring architectural fidelity for performance-critical ARM targets.

 \subsection{Training}
 All hyperparameter optimization experiments were conducted on a small 500k portion of AnghaBench. We tested various hyperparameter settings on this subset of our benchmark. After identifying the optimal configuration, we scaled up the training data to 1.31M samples. We trained three models: \texttt{DeepSeek-Coder1.3B}~\cite{deepseek-coder}, \texttt{Qwen2.5-Coder} (1.5B and 0.5B)~\cite{qwen25-coder}. Given the dataset size of 1.3M million samples, with an average of 13k tokens per sample, we opted for smaller models. Training was done on A100 GPUs (40GB each). Training with 1.3M samples, a batch size of 24, and 2 epochs required three days. To conserve memory, mixed precision training with bfloat16 was employed. Given limited capacity for large batch sizes, we applied gradient accumulation with an effective batch size of 2. We used paged AdamW~\cite{adamw} to avoid memory spikes, with a weight decay of 0.001. We chose a small learning rate of \( 2 \times 10^{-5} \)with a cosine schedule, as experiments indicated this schedule performed best. We trained our model with a context window of 16k.  In inference, we do RoPE~\cite{rope} extrapolation to increase the context window to 32.7k.

 \begin{table}[h]
    \centering
    \small
    \renewcommand{\arraystretch}{1.3}
    \begin{tabular}{l l}
        \toprule
        \rowcolor{gray!15}
        \textbf{Input} & \textbf{\texttt{ldr r1, r2}} \\
        \midrule
        \rowcolor{gray!15}
        \textbf{Tokenizer} & \textbf{Tokens} \\
        \midrule
        DeepSeek/Qwen 2.5 coder & 
        \begin{tabular}[t]{@{}l@{}}
            \colorbox{red!15}{\strut\texttt{ld}}%
            \colorbox{yellow!15}{\strut\texttt{r}}%
            \colorbox{green!15}{\strut\texttt{\char32}}%
            \colorbox{blue!15}{\strut\texttt{r}}%
            \colorbox{yellow!15}{\strut\texttt{1}}%
            \colorbox{orange!15}{\strut\texttt{,}}%
            \colorbox{purple!15}{\strut\texttt{\char32}}%
            \colorbox{cyan!15}{\strut\texttt{r}}%
            \colorbox{yellow!15}{\strut\texttt{2}}
        \end{tabular} \\[2ex]
        \GGn Extended Tokenizer & 
        \begin{tabular}[t]{@{}l@{}}
            \colorbox{red!15}{\strut\texttt{ldr}}%
            \colorbox{green!15}{\strut\texttt{\char32}}%
            \colorbox{blue!15}{\strut\texttt{r1}}%
            \colorbox{orange!15}{\strut\texttt{,}}%
            \colorbox{purple!15}{\strut\texttt{\char32}}%
            \colorbox{cyan!15}{\strut\texttt{r2}}
        \end{tabular} \\
        \bottomrule
    \end{tabular}
    \caption{Comparison of tokenization approaches between DeepSeek/Qwen-Coder and our extended tokenizer. Spaces are represented as \texttt{\char32} and shown with colored backgrounds to highlight token boundaries. Note how our tokenizer groups related tokens (e.g., \texttt{ldr} and \texttt{r1}) as singular units.}
    \label{tab:tokenizer_comparison}
\end{table}

\subsection{Tokenizer Extension} 

To improve our LLMs' capability in comprehending and generating assembly code, we augmented the tokenizer by incorporating the most common opcodes and register names from x86 and ARMv5/ARMv8/RISC-V64 architectures (as shown in Table~\ref{tab:tokenizer_comparison}). This targeted design improves token alignment with instruction semantics, enabling more precise and efficient assembly translation. As shown in table~\ref{tab:fertility_rate}, our extension decreases the fertility rate (tokens/words)~\cite{fertility} of Qwen and Deepseek tokenizers by 2.65\% and 6.9\%, respectively. This corresponds to our model fitting 848 and 2.2k tokens respectively.

\begin{table}[ht]
\centering
\resizebox{\linewidth}{!}{
\begin{tabular}{lcccc}
\toprule
\textbf{Model} & \textbf{x86} & \textbf{ARMv5} & \textbf{ARMv8} & \textbf{RISC-V64} \\
\midrule
Qwen-Coder~\cite{qwen2.5-coder}        & 4.28 & 2.89 & 3.62 & 3.62 \\
DeepSeek-Coder~\cite{deepseek-coder}    & 3.74 & 3.51 & 4.28 & 4.28 \\
\GGn-Qwen (Ours)      & \textbf{4.14} & \textbf{2.87} & \textbf{3.50} & \textbf{3.50} \\
\GGn-DeepSeek (Ours)      & \textbf{3.47} & \textbf{3.26} & \textbf{3.99} & \textbf{3.37} \\
\midrule
\rowcolor{blue!10}
$\Delta$ Qwen (\%)   & $\downarrow$3.3\% & $\downarrow$0.5\% & $\downarrow$3.4\% & $\downarrow$3.4\% \\
\rowcolor{blue!10}
$\Delta$ DeepSeek (\%)   & $\downarrow$7.2\% & $\downarrow$6.9\% & $\downarrow$6.8\% & $\downarrow$6.8\% \\
\bottomrule
\end{tabular}
}
\caption{Tokenizer fertility rate (tokens/words) across ISAs. Lower is better.}
\label{tab:fertility_rate}
\end{table}

\begin{table*}[t]
\centering
\resizebox{\textwidth}{!}{%
\begin{tabular}{|c||cc|cc|cc|}
\hline
\rowcolor{pink!20}
{\textbf{Model}} 
& \multicolumn{2}{c|}{\textbf{ARMv5}} 
& \multicolumn{2}{c|}{\textbf{ARMv8}} 
& \multicolumn{2}{c|}{\textbf{ARMv8}} \\
\rowcolor{pink!20}
& \textbf{HumanEval} & \textbf{HumanEval} 
& \textbf{HumanEval} & \textbf{HumanEval} 
& \textbf{BringupBench} & \textbf{BringupBench} \\
\hline
& \textbf{-O0} & \textbf{-O2} & \textbf{-O0} & \textbf{-O2} & \textbf{-O0} & \textbf{-O2} \\
\hline
GPT-4o~\cite{gpt4o} & 8.48\% & 3.64\% & 10.3\% & 4.24\% & 1.54\% & 0\% \\
Qwen2.5-Coder-1.5B~\cite{qwen2.5-coder}  & 0\%  & 0\%  & 0\% & 0\% & 0\% & 0\% \\
Qwen2.5-Coder-3B~\cite{qwen2.5-coder} & 0.61\% & 0\%  & 0\%  & 0\% & 0\% & 0\% \\
StarCoder2-3B~\cite{starcoder}  & 0\% & 0\% & 0\% & 0\% & 0\% & 0\% \\
Deepseek-R1-1.5B~\cite{deepseek-r1}  & 0\% & 0\% & 0\% & 0\% & 0\% & 0\% \\
Deepseek-R1-Qwen-7B~\cite{deepseek-r1}  & 0\% & 0\% & 0\% & 0\% & 0\% & 0\% \\
\hline
\rowcolor{blue!5}
\GGn-Deepseek-1.3B  & 79.25\% & 12.80\% & 75.15\% & 10.3\% & 3.08\% & 0\% \\
\rowcolor{blue!10}
\GGn-0.5B  & 90.85\% & 23.03\% & 86.06\% & 25.45\% & 27.69\% & 3.08\% \\
\rowcolor{blue!15}
\GGn-1.5B  & \textbf{93.71\%} & \textbf{50.30\%} & \textbf{99.39\%} & \textbf{45.12\%} & \textbf{49.23\%} & \textbf{15.38\%} \\
\hline
\end{tabular}
}
\caption{
% Performance across benchmarks and optimization levels. Best results are highlighted.
Models trained with our method outperform baselines across all benchmarks, at all optimization levels.
}

\label{tab:results}
\end{table*}

\section{Experiments and Evaluation}
\label{sec:eval}

In this section, we describe our experimental setup, training methodology, evaluation benchmarks, and the metrics used to assess the accuracy and robustness of our CISC-to-RISC transpiler.

\subsection{Setup}
 We leveraged LLaMa-Factory~\cite{llamafac}, DeepSpeed Zero3~\cite{deepspeed}, liger kernels~\cite{liger-kernel}, and FlashAttention2~\cite{flashattn} for efficient training and memory optimization. We also used caching to enhance inference speed and disabled sampling to ensure deterministic outputs. We used vLLM~\cite{vllm} to deploy our model and achieve a throughput of 36x requests per second at 32.7k tokens context window on a single A100 40GB GPU. Additionally, We apply post-quantization using \texttt{llama.cpp}~\cite{llama-cpp} (e.g., \texttt{bfloat16}, \texttt{int8}, \texttt{int4}) to optimize inference for CPU-based deployment.

\subsection{Evaluation}\label{sec:metrics}

We evaluate \GG using two complementary benchmarks: HumanEval-C~\cite{tan2024llm4decompile} and BringUpBench~\cite{austin2024bringup}. HumanEval was originally introduced by \citet{humaneval} for Python code generation. The benchmark consists of 164 programming problems that assess language comprehension, reasoning, and algorithmic thinking. For our evaluation, we utilize the C-translated version from LLM4Decompile \cite{tan2024llm4decompile}, which maintains the same problems while converting both function implementations and test cases to C code. 

\begin{figure}[h]
    \centering
    \includegraphics[width=\linewidth]{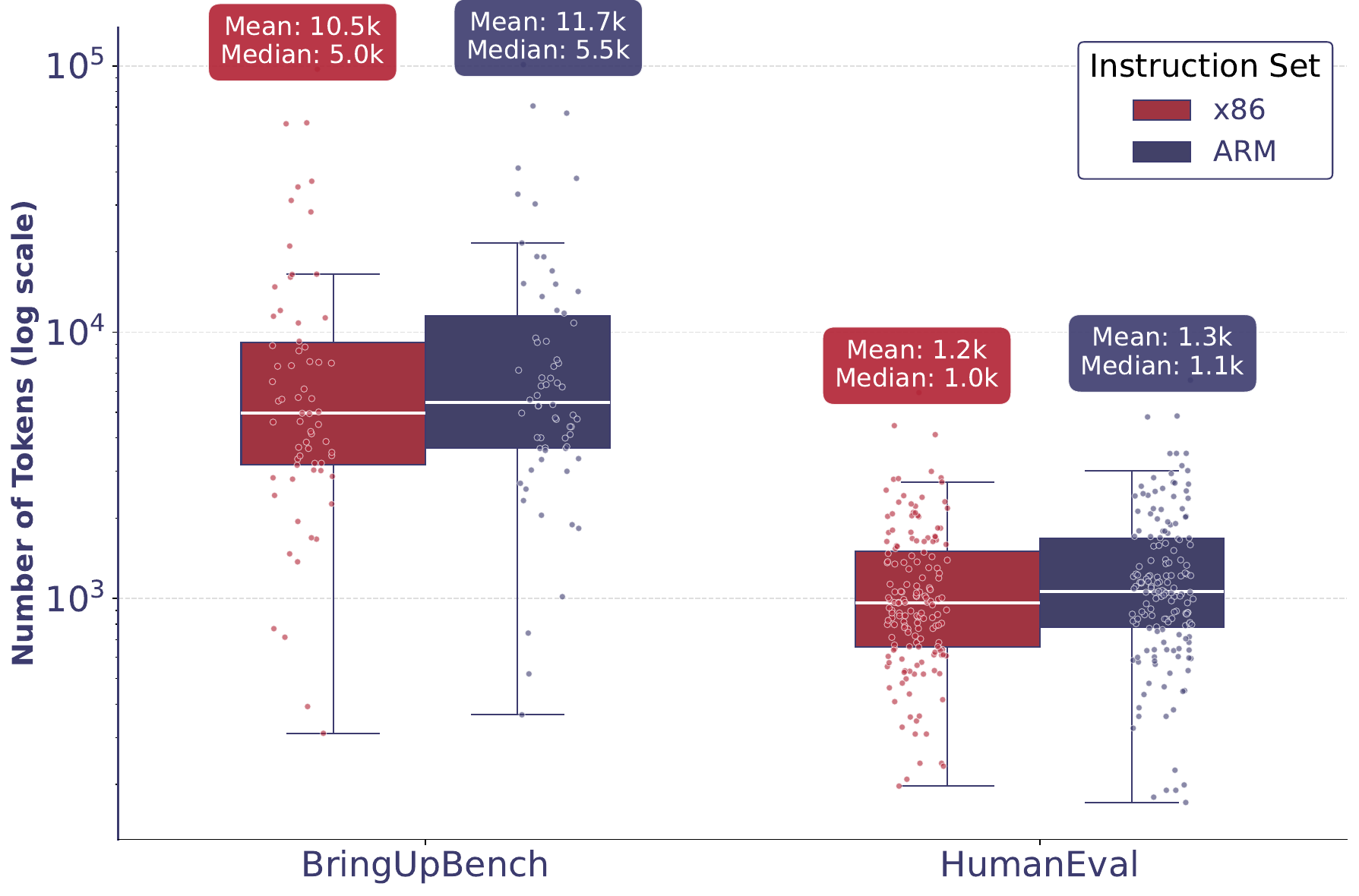}
    \caption{Token counts by ISA and benchmark; BringUpBench is substantially longer than HumanEval.}
    \label{fig:token-distribution}
\end{figure}

To evaluate real-world generalization, we leverage BringUpBench~\cite{austin2024bringup}, a challenging benchmark of 65 bare-metal programs ranging from 85 to 5751 lines of code. Unlike HumanEval, which consists of isolated functions, BringUpBench programs are embedded in full project structures with internal libraries and cross-linked components. This setup more accurately reflects real-world embedded systems development, where executing even a single file often requires compiling and linking the entire codebase. As a result, BringUpBench imposes significantly greater context length demands. On average, each BringUpBench sample requires 8.9× more tokens for x86 and 8.8× more for ARM compared to HumanEval, as shown in Figure~\ref{fig:token-distribution}. The benchmark’s diverse control flow and I/O patterns further elevate its difficulty, making it a strong testbed for assessing the robustness and scalability of our transpiler.

We use \texttt{gcov}, GNU's coverage tool, to measure line coverage, a core metric in software testing that captures which code lines were executed at least once, thereby exposing untested paths and blind spots~\cite{myers2011art}. HumanEval and BringupBench achieved 98.81\% and 97.32\% average coverage, respectively, indicating near-complete execution of all code lines during testing.

We evaluate functional correctness by executing the transpiled ARM code against full unit test suites. A prediction is deemed correct only if all test cases pass, partial correctness is not counted. For HumanEval, this involves compiling the predicted code, linking it with the provided tests, and executing the binary as shown inf figure~\ref{fig:transpiler-overview}. For BringUpBench, we leverage its Makefile to build the static library and link it with the target file. The output is then compared against the expected output using a diff-based check. This strict pass@1 evaluation, based solely on the most probable sample, even when beam search (beam size = 8) is used, ensures that only fully functional translations contribute to final accuracy.

\section{Results and Analysis}\label{sec:results}

We evaluate the efficacy of our transpiler for CISC-to-RISC assembly translation, focusing on the correctness of the output ARM assembly. Utilizing the metrics defined above (\S\ref{sec:eval}), we compare our approach with state-of-the-art coding LLMs and evaluate our approach for x86 to ARM transpilation (Table\ref{tab:results}).

\subsection{Transpiler Validation}

\paragraph{Baselines.} As shown in Table~\ref{tab:results}, most baseline models, including state-of-the-art LLMs such as StarCoder2~\cite{starcoder}, DeepSeek~\cite{deepseek-coder}, and Qwen2.5~\cite{qwen2.5-coder}, achieve 0\% accuracy in all transpilation tasks, underscoring the unique difficulty of low-level ISA translation. These models, while effective on high-level programming benchmarks, lack the architectural grounding and token-level inductive bias needed to generalize from x86 to ARM. GPT-4o was the only exception, achieving 1.5-8\% accuracy, which remains far below usable thresholds, highlighting that general-purpose LLMs are not yet suitable for assembly-level translation without specialized training. This performance gap reinforces the need for task-specific instruction tuning and architectural adaptation to handle the deep structural mismatch between CISC and RISC.

\paragraph{\GG Results.} Our \GG models, particularly the \GGn-1.5B variant, substantially outperform all baselines, reaching 99.39\% accuracy on ARMv8 and 93.71\% on ARMv5 under the \texttt{-O0} setting. This validates the effectiveness of architecture aware training, tokenizer extension, and longer context modeling in capturing fine-grained register and memory semantics. For \texttt{-O2} optimized code, accuracy drops to 45.12\% (ARMv8) and 50.30\% (ARMv5), exposing the fragility of current LLMs under aggressive compiler transformations. This suggests that while our model learns to generalize well under minimal optimization, it struggles with control/data flow reordering and register coalescing introduced by \texttt{-O2} passes. Addressing this challenge may require incorporating optimization-invariant representations, such as symbolic traces or control/data-flow graphs, or extending the training set with more aggressively optimized samples.A detailed error analysis can be found in Appendix~\ref{sec:scale_and_error}.

\paragraph{RISC-v64.} To demonstrate the generality of our method, we also trained our model on the task of transpiling from x86 to RISC-V64, achieving a pass@1 of 89.63\%. Notably, our model significantly outperforms existing models like GPT4o and DeepSeekCoder2-16B, which achieved much lower test accuracies of 7.55\% and 6.29\%, respectively. This result is 9\% lower than ARMv8 which shows how much different RISC-v64 from x86 compared ARMv8.

\paragraph{\texttt(-O2) Opt.} Compiler optimizations (-O2) introduce complex patterns that increase failure frequency compared to \texttt{-O0}. A common error is the motion of the instruction; for example, misplacing \texttt{ cbz}\footnote{Compare and Branch if Zero} alters the control flow, revealing the difficulty of the model in interpreting optimized sequences. While hard to detect automatically, such errors can be repaired via manual inspection~\citep{liu2025craftrtl}, symbolic solvers~\citep{lee2024guess,mora2024synthetic}, or reasoning models. Hybrid human-AI approaches may improve correctness guarantees.

\begin{table}[t]
\centering
\scriptsize
\begin{minipage}{0.48\textwidth}
\resizebox{\textwidth}{!}{%
\begin{tabular}{|p{2.5cm}|p{3.89cm}|}
\hline
\textbf{Error Type} & \textbf{Files with Errors after Guess} \\
\hline
Input + output out of context window & LongDiv, Regex-Parser, RLE-Compress, FFT-Int, Blake2B, Anagram, C-Interp, Totient,  Banner,  Lz Compress, Satomi, Rho-Factory \\ \hline
Duplicate function error & Frac-Calc, Minspan \\
\hline
Stack/memory error & Boyer-Moore-Search, Topo-Sort, Audio-Codec,  Weekday, Simple-Grep, Max-Subseq,  Priority-Queue, Dhrystone, Cipher, AVL-Tree, QSort-Demo, Vectors-3D, Pascal \\
\hline
Missing function error & Fuzzy-Match, Tiny-NN, Kadane, Audio-Codec, Frac-Calc, Kepler, Dhrystone, Cipher,  Graph-Tests, Quaternions, AVL-Tree, K-Means, QSort-Demo, Vectors-3D \\
\hline
Labels referred but not defined & Fuzzy-Match, Life, AVL-Tree, K-Means \\
\hline
Register mislabel error & Bloom-Filter, Topo-Sort, Weekday,  Knights-Tour, Simple-Grep, Max-Subseq,  Mersenne, Audio-Codec, K-Means, QSort-Demo,  Vectors-3D, Pascal, Minspan \\
\hline
Incorrect immediate value & Kadane \\
\hline
\end{tabular}}
\caption{
Failed files on BringupBench. Errors after the Guess stage are largely around dataflow reasoning. File names are grouped by error type.
}
\label{tab:error_types_guess_only}
\end{minipage}
\end{table}

\paragraph{BringUpBench.
} We evaluate \GGn-1.5B on BringUpBench~\cite{austin2024bringup} and manually analyze over 200 unit-tested binaries. Our model achieves 49.23\% exact match accuracy under \texttt{-O0} (Table~\ref{tab:results}) with virtually no syntax errors, outputs consistently adhere to valid ARM assembly with correct opcodes, registers, and memory access. This reflects a strong surface-form prior, shifting focus to semantic errors like incorrect dataflow. Notably, 17\% of failures stem from context truncation, indicating a key limitation of current context window sizes. Table~\ref{tab:error_types_guess_only} summarizes common failure types, including duplicate code, invalid control flow, misused registers / intermediaries, and stack errors - most symptomatic of broken data flow rather than syntax issues. These may be alleviated through longer training, symbolic repair, or richer representations. Lastly, the benchmark’s extensive unit tests offer a valuable semantic signal in the absence of ground truth, suggesting a compelling path for test-driven transpilation and iterative repair.

\begin{figure*}[!htbp]
    \centering
    \includegraphics[width=\linewidth]{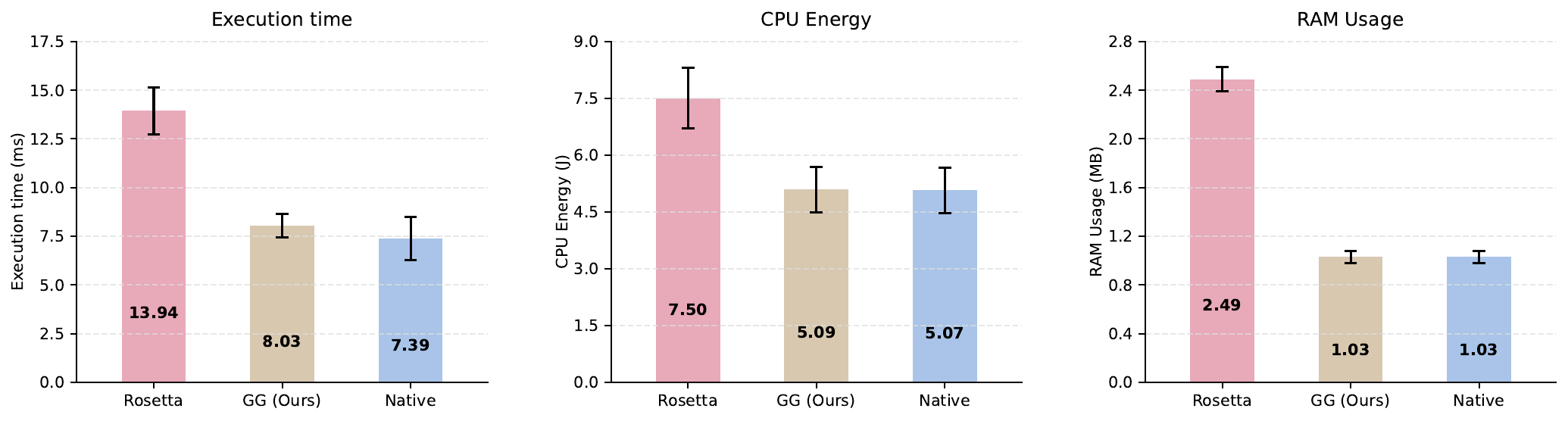}
    \caption{Comparison of execution time, energy consumption, and memory usage across Rosetta, \GGn, and native binaries.}

    \label{fig:performance}
\end{figure*}

\subsection{Real-World Case Study}
To evaluate the efficiency of our transpiler, we conducted a real-world study on an Apple M2 Pro (ARM64v8-A). This setup offers two advantages: (1) native ARM toolchain support, avoiding cross-compilation; and (2) Apple’s Rosetta 2 layer, enabling consistent evaluation across execution modes on the same hardware. We assess performance across three environments: (i) native ARM64 binaries, (ii) x86 binaries via Rosetta 2, and (iii) \GGn-transpiled x86-to-ARM64 assembly. For each, we measure execution time, CPU energy (via \texttt{powermetrics}), and memory usage. Each program is executed 100 times, reporting the geometric mean~\cite{fleming1986not}, under controlled conditions.

\noindent Figure~\ref{fig:performance} shows that \GG achieves near-native performance: matching execution time, 1.73$\times$ faster than Rosetta, with 1.47$\times$ better energy efficiency and 2.41$\times$ better memory usage. \GGn’s memory footprint (1.034 MB) is nearly identical to native (1.03 MB), while Rosetta uses 2.49 MB.

These results demonstrate that LLM-based binary translation offers a compelling alternative to traditional dynamic translation layers like Rosetta. Unlike Rosetta, which incurs a persistent runtime overhead, \GG performs a one-time transpilation, avoiding the cumulative “runtime tax” and enabling leaner, faster execution. Moreover, our approach is general-purpose and untethered to Apple’s ecosystem, enabling broader cross-ISA deployment and efficient CISC-to-RISC translation across diverse platforms. See Appendix~\ref{sec:scale_and_error} for scaling, quantization, and error analysis.

\begin{figure*}[!htbp]
    \centering
    \begin{subfigure}{0.52\linewidth}
        \centering
        \includegraphics[width=\linewidth]{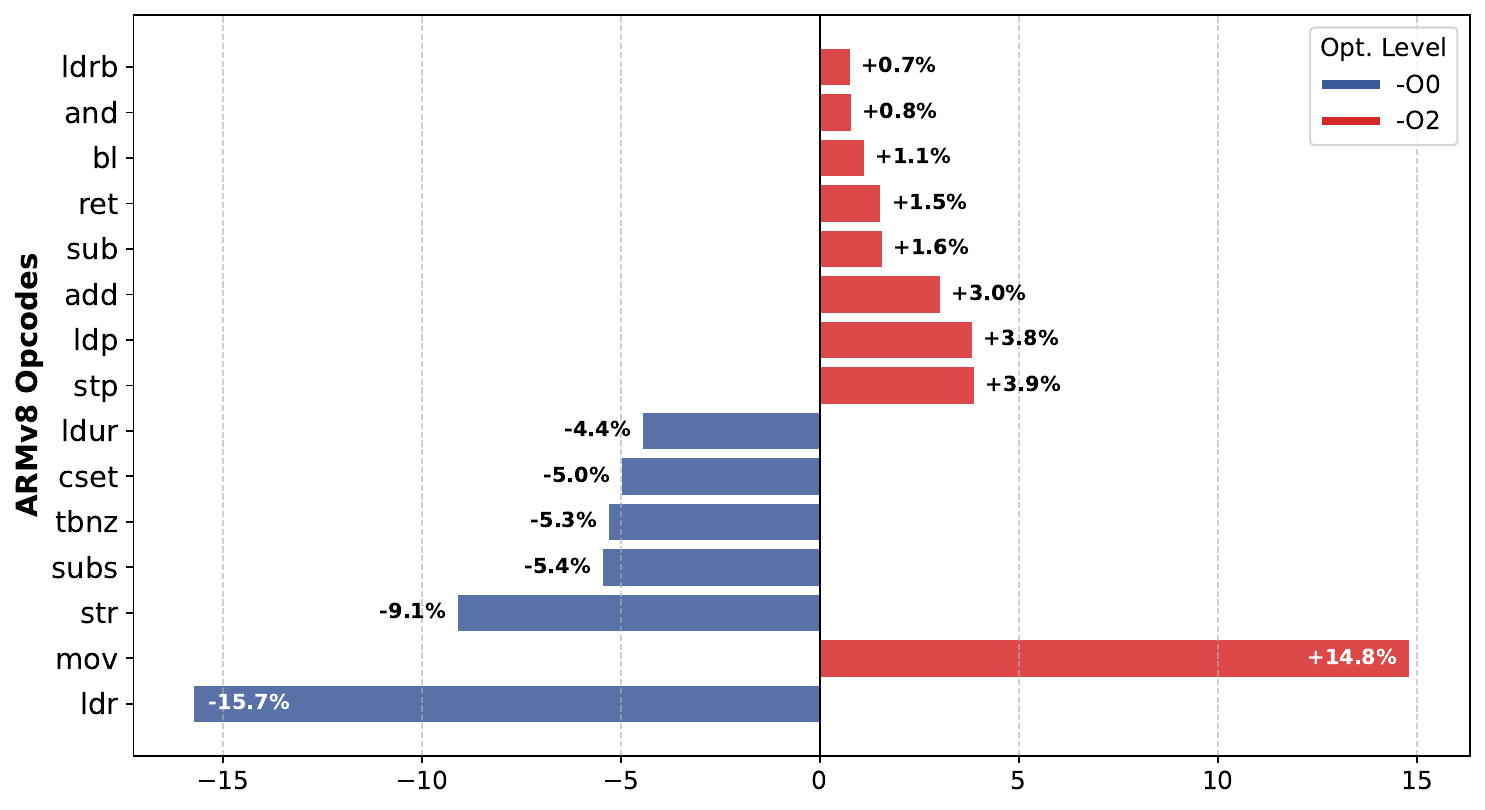}
        \caption{Opcode shift distribution in ARMv8}
        \label{fig:asm-armv8-shift}
    \end{subfigure}
    \hfill
    \begin{subfigure}{0.46\linewidth}
        \centering
        \includegraphics[width=\linewidth]{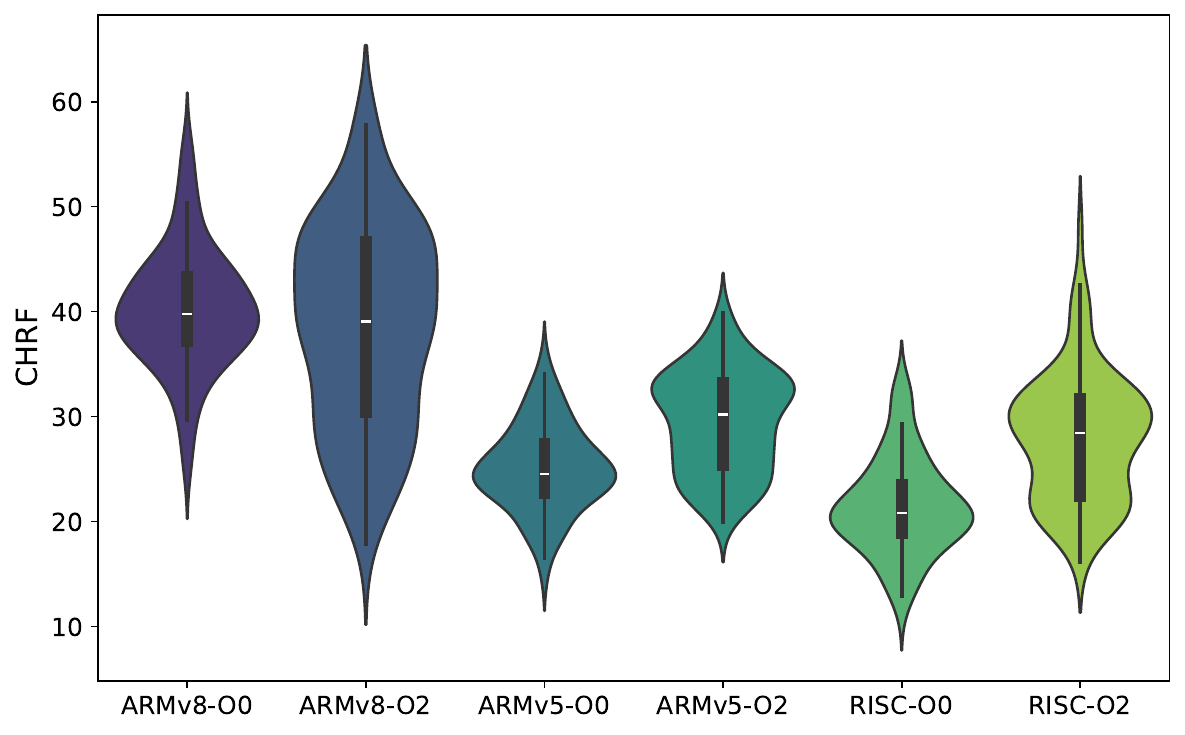}
        \caption{CHRF similarity scores}
        \label{fig:asm-chrf-sim}
    \end{subfigure}
    \caption{Side-by-side comparison of opcode shift and CHRF similarity in ARM assembly analysis.}
    \label{fig:asm-side-by-side}
\end{figure*}

\subsection{Similarity Analysis Across ISAs}
In Figure~\ref{fig:asm-chrf-sim}, we observe that ARMv8 exhibits the highest average similarity to x86 (40.19\%), followed by ARMv5 (25.09\%) and RISC-V64 (21.41\%). This gradient of similarity directly correlates with the drop in model accuracy from ARMv8 (99.39\%) to ARMv5 (93.71\%) and further down to RISC-V (89.63\%). We hypothesize that this discrepancy is rooted in the increasing divergence in instruction semantics and register abstractions across these ISAs. ARMv8’s shift toward CISC-like design~\cite{redhat_arm_vs_x86} likely boosts its alignment with x86, aiding model generalization. In contrast, ARMv5 and RISC-V have simpler, more divergent instruction sets and addressing schemes, making the x86-to-RISC mapping less predictable and thus harder to learn.

Figure~\ref{fig:asm-armv8-shift} highlights a significant shift in ARMv8 opcode usage between \texttt{-O0} and \texttt{-O2}. At \texttt{-O2}, \texttt{mov} becomes dominant (+14.8\%), indicating more register reuse and reduced memory traffic via explicit \texttt{ldr}/\texttt{str}. This hides direct data movement, making it harder for the model to learn memory interaction. Paired instructions like \texttt{ldp}/\texttt{stp} appear more frequently, packing semantics into fewer lines, while conditional ops (\texttt{tbnz}, \texttt{cset}) are folded into predicated sequences. These changes, introduced by the compiler, abstract both control and data flow. We hypothesize that the model, trained only on \texttt{-O2}, must decode complex x86 semantics into a highly optimized and compressed ARMv8 form. This transformation increases learning difficulty and explains the drop in \texttt{-O2} accuracy (to 45.12\%) despite strong \texttt{-O0} performance.

\begin{table}[htbp]
\centering
\renewcommand{\arraystretch}{1.15}
\setlength{\tabcolsep}{8pt}
\sffamily
\resizebox{\linewidth}{!}{%
\begin{tabular}{lcc}
\hline
\textbf{Model Variant} & \textbf{ARMv8 Accuracy} & \textbf{Impact (\(\Delta\))} \\
\hline
Qwen2.5-Coder & 0\% & -- \\
\rowcolor{blue!2}
+ 1M AnghaBench & 93.94\% & \(+93.94\%\) \\
\rowcolor{blue!4}
+ 0.3M Stackv2 & 95.38\% & \(+1.44\%\) \\
\rowcolor{blue!6}
+ RoPE Extrapolation & 97.14\% & \(+1.76\%\) \\
\rowcolor{blue!8}
+ Extended Tokenizer & 98.18\% & \(+1.04\%\) \\
\rowcolor{blue!10}
+ 8 Beam Search  & \textbf{99.39\%} & \(\mathbf{+1.21\%}\) \\
\hline
\end{tabular}
}
\caption{Ablation study showing incremental improvements on ARMv8 accuracy from each added component.}
\label{tab:ablation}
\end{table}

\subsection{Ablation Study}
To understand what contributed most to model performance, we performed ablations shown in Table~\ref{tab:ablation}, focusing on four key aspects: training data size, RoPE extrapolation, the extended tokenizer, and decoding strategy.

First is the training data. As we increased the amount of training data to 1M AnghaBench, the accuracy jumps from 0\% to 93.94\%; including an additional 0.3M Stackv2 data points further improves accuracy to 95.38\%. While effective, this scaling approach depends on high-quality, large-scale datasets and longer training time. Second is the architectural enhancement through RoPE Extrapolation, which pushes performance to 97.14\%, indicating a +1.76\% improvement. This suggests that enabling better generalization beyond the fixed context window substantially benefits instruction understanding and long-range dependency modeling.

The third contributing factor is tokenizer coverage: by extending the tokenizer to include additional subword units and symbols, we observe a further gain to 98.18\%, adding +1.04\%, highlighting the importance of adapting the tokenizer to the domain-specific vocabulary of assembly code. Finally, decoding strategy plays a non-trivial role; switching to 8-beam search yields the final boost to 99.39\%, adding another +1.21\%. Altogether, this progression shows that while data scaling gives the biggest leap, fine architectural and decoding choices compound gains toward near-perfect accuracy.

\section{Conclusion}\label{sec:conc}

We introduce \textit{Guaranteed Guess} (\GG), a language-model-based CISC-to-RISC transpiler that unifies pre-trained LLMs with a test-driven validation framework. \GG directly transpiles x86 assembly into efficient ARM and RISC-V binaries while embedding unit tests to enforce functional correctness. Through architectural enhancements, such as tokenizer extension, RoPE extrapolation, and beam decoding, \GG achieves 99. 39\% accuracy in HumanEval and 49. 23\% in BringUpBench, outperforming both strong LLMs and dynamic virtualization systems like Rosetta. Our analysis highlights how ISA similarity and compiler optimizations affect accuracy, with \GG achieving 1.73× faster execution, 1.47× lower energy use, and 2.41× smaller memory footprint than Rosetta on real-world binaries. These results position \GG as a scalable, test-verified solution for efficient, cross-ISA binary translation.

\section{Limitations}

While \textit{Guaranteed Guess} presents a significant advancement in CISC-to-RISC transpilation using LLMs, several limitations remain. First, the model's performance degrades substantially under compiler optimization flags (e.g., \texttt{-O2}), highlighting its sensitivity to code transformation patterns that abstract data and control flow. This suggests a need for stronger semantic modeling or auxiliary representations such as control/data-flow graphs. Second, the ``guarantee'' provided by \GG is inherently bounded by the quality and coverage of the unit tests. While unit test success is a strong functional proxy, it cannot ensure full semantic equivalence or optimality of the transpilation. Lastly, the evaluation excludes compiler-, symbolic-, or heuristic-based transpilation baselines, leaving open questions about hybrid system effectiveness and competitive upper bounds.

\bibliography{references}

\newpage
\appendix

\section{Appendix}

\subsection{Extra Data Analysis}
\label{sec:scale_and_error}

\begin{figure*}[t]
    \centering
    \small
    \includegraphics[width=0.9\linewidth]{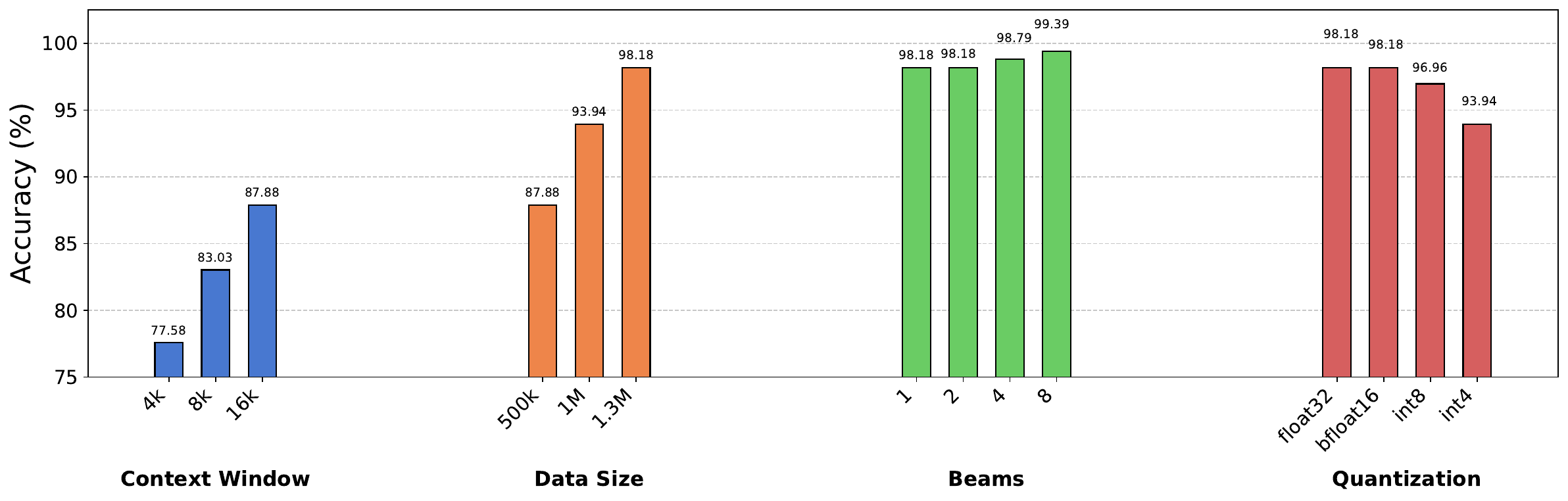}
    \caption{\textbf{Impact of scaling and quantization on Qwen2.5-Coder 1.5B variant evaluated using the \textit{code coverage} metric on HumanEval with -O0 compiler optimization.}}
    \label{fig:comparision}
\end{figure*}

\paragraph{Scaling and quantization effect on Qwen2.5-coder models.} Figure~\ref{fig:comparision} represents an study to understand where most of the training benefit for our transpiler originates. In particular, we focus on three fundamental modeling aspects and describe their impact on the asm-to-asm transpiler.

Our first and most significant result relates to the context window size, and its impact on the transpiler. Recall that a model's context window is the amount of text, in tokens, that the model can consider or “remember” at any one time. We found that programs do not fully fit in the context window (which includes both the input and output of the model, i.e., the x86 asm and the generated ARM asm), are very likely to not pass all our tests. Increasing the context window length during training had a big impact on our model's accuracy, where going from 4k to 16k improved the total number of fully correct transpiled programs by 10\% points, roughly an additional 16 programs out of the 164 total in HumanEval.

\begin{table}[h]
\large
\centering

\resizebox{\linewidth}{!}{
\begin{tabular}{@{}p{0.8cm}p{0.8cm}p{9.9cm}@{}}
\toprule
\textbf{Prog ID} & \textbf{Edit Dist} & \textbf{Example} \\
\midrule
P37 & 1 & \textbf{Incorrect immediate value causes wrong division factor and early loop termination} \\
& & Ground truth: \code{asr} \reg{r2}, \reg{r2}, \imm{\#2} \\
& & Predicted: \code{asr} \reg{r2}, \reg{r2}, \imm{\#1} \\
\midrule
P127 & 1 & \textbf{Array index offset error causes wrong element comparison} \\
& & Ground truth: \code{sub} \reg{r3}, \reg{r3}, \imm{\#2} \\
& & Predicted: \code{sub} \reg{r3}, \reg{r3}, \imm{\#1} \\
\midrule
P63 & 12 & \textbf{Register overwrite corrupts loop counter before multiplication} \\
& & Ground truth: \code{mov} \reg{r0}, \reg{r2}; \code{ldr} \reg{r1}, \mem{[r3, r1, lsl \#2]}; \code{mul} \reg{r0}, \reg{r0}, \reg{r1} \\
& & Predicted: \code{ldr} \reg{r0}, \mem{[r3, r1, lsl \#2]}; \code{mul} \reg{r0}, \reg{r0}, \reg{r1} \\
\midrule
P153 & 17 & \textbf{Incorrect instruction sequence fails to compute absolute value} \\
& & Ground truth: \code{sub} \reg{r2}, \reg{r2}, \reg{r3}; \code{cmp} \reg{r2}, \imm{\#0}; \code{rsblt} \reg{r2}, \reg{r2}, \imm{\#0} \\
& & Predicted: \code{sub} \reg{r1}, \reg{r2}, \reg{r3}; \code{eor} \reg{r2}, \reg{r1}, \reg{r2}; \code{sub} \reg{r2}, \reg{r2}, \reg{r1} \\
\midrule
P47 & 19 & \textbf{Mismatched memory access offsets cause incorrect data retrieval} \\
& & Ground truth: \code{str} \reg{r1}, \mem{[fp, \#-404]}; \code{ldr} \reg{r2}, \mem{[fp, \#-404]} \\
& & Predicted: \code{str} \reg{r1}, \mem{[fp, \#-404]}; \code{ldr} \reg{r2}, \mem{[r3, \#-20]} \\
\bottomrule
\end{tabular}}
\caption{Armv5 Syntactically similar generations can still produce critical semantic errors. 
}
\label{tab:critical_errors}
\end{table}

The second effect of scaling we observed and leveraged was that training on more data also played a major role in our transpiler's efficacy. As shown in Figure~\ref{fig:comparision}, using a context window of 16k and increasing the training data from 500k samples to 1.3 million samples further increased and pushed the accuracy up to about 98\% from 87\%. This is generally a challenging method of scaling, as obtaining more data with good quality is not always available and also results in increased total training time of the model. 

The third scaling impact we found was the benefit of increasing the number of beams and doing a beam search. Beam search is a heuristic search algorithm which allows the model to explore multiple token paths
in parallel during an inference. Intuitively, beam search allows the model to explore alternative options for next token generation, settling on the most likely token. Beam searching presents an obvious trade-off between computational resources utilization for an inference and prediction accuracy. Combined with a large context window, this is a very powerful technique which we found to be more pronounced when a model was not already near perfect accuracy: in Figure~\ref{fig:comparision}, we show an increase going up to 99.39\%
with the use of beam search for assembly transpilation. We found diminishing returns for using more than 4 beams on accuracy.

Finally, from an efficiency perspective, we show that aggressive quantization does not severely impact our transpilers accuracy. Going from FP32 down to INT4 substantially reduces the transpilers inference footprint, with a minimal (less than 4\%) impact on model prediction accuracy. This shows the potential of designing small enough models for deployment on edge devices, which we would envision the GG transpiler to
be used for CISC-to-RISC translations in practice.

\noindent \paragraph{Transpilation Error Analysis.} \label{transpilation_error} 
We provide a detailed analysis of functionally equivalent predictions produced by our model that deviate syntactically from the ground truth. Such cases reveal the model's ability to generalize instruction patterns while maintaining semantic correctness, a desirable trait in low-level code generation where multiple implementations can achieve the same functional outcome.

\begin{table}[h]
\large
\centering

\resizebox{\linewidth}{!}{\begin{tabular}{@{}p{0.8cm}p{0.8cm}p{9.9cm}@{}}
\toprule
\textbf{Prog ID} & \textbf{Edit Dist} & \textbf{Example} \\
\midrule
P108 & 16 & \textbf{Different registers can be chosen for temporary values while maintaining same data flow} \\
& & Ground truth: \code{mov} \reg{r2}, \reg{r0}; \code{add} \reg{r2}, \reg{r2}, \imm{\#1} \\
& & Predicted: \code{mov} \reg{r3}, \reg{r0}; \code{add} \reg{r3}, \reg{r3}, \imm{\#1} \\
\midrule
P8 & 12 & \textbf{Local variables can be stored at different stack locations while maintaining correct access patterns} \\
& & Ground truth: \code{str} \reg{r1}, \mem{[fp, \#-8]}; \code{str} \reg{r2}, \mem{[fp, \#-12]} \\
& & Predicted: \code{str} \reg{r1}, \mem{[fp, \#-12]}; \code{str} \reg{r2}, \mem{[fp, \#-8]} \\
\midrule
P119 & 6 & \textbf{Compiler-generated symbol names can differ while referring to same data} \\
& & Ground truth: \code{.word} \code{out.4781} \\
& & Predicted: \code{.word} \code{out.4280} \\
\midrule
P135 & 12 & \textbf{Multiple instructions can be combined into single equivalent instruction} \\
& & Ground truth: \code{mov} \reg{r3}, \reg{r0}; \\
& & \hspace{2.50cm}\code{str} \reg{r3}, \mem{[fp, \#-8]} \\
& & Predicted: \code{str} \reg{r0}, \mem{[fp, \#-8]} \\
\midrule
P162 & 4 & \textbf{Stack frame offsets can vary while maintaining correct variable access} \\
& & Ground truth: \code{strb} \reg{r3}, \mem{[fp, \#-21]} \\
& & Predicted: \code{strb} \reg{r3}, \mem{[fp, \#-17]} \\
\midrule
P88 & 23 & \textbf{Memory allocation sizes can vary if sufficient for program needs} \\
& & Ground truth: \code{mov} \reg{r0}, \imm{\#400} \\
& & Predicted: \code{mov} \reg{r0}, \imm{\#800} \\
\midrule
P103 & 52 & \textbf{Different instruction sequences can achieve same logical result} \\
& & Ground truth: \code{cmp} \reg{r3}, \imm{\#0}; \code{and} \reg{r3}, \reg{r3}, \imm{\#1}; \code{rsblt} \reg{r3}, \reg{r3}, \imm{\#0} \\
& & Predicted: \code{rsbs} \reg{r2}, \reg{r3}, \imm{\#0}; \code{and} \reg{r3}, \reg{r3}, \imm{\#1}; \code{and} \reg{r2}, \reg{r2}, \imm{\#1}; \code{rsbpl} \reg{r3}, \reg{r2}, \imm{\#0} \\
\midrule
P69 & 50 & \textbf{Constants can be loaded directly or from literal pool} \\
& & Ground truth: \code{mvn} \reg{r3}, \imm{\#-2147483648} \\
& & Predicted:\newline \code{ldr} \reg{r3}, \imm{.L8}; \code{.L8:} \code{.word} \imm{2147483647} \\
\bottomrule
\end{tabular}}
\caption{Simple Variation Patterns in Functionally Equivalent Code}
\label{tab:simple_variations}
\end{table}

Table~\ref{tab:simple_variations} enumerates a range of examples with moderate edit distances, where syntactic differences arise from register allocation, operand ordering, and memory layout choices. For instance, the model often selects different temporary registers (e.g., \texttt{r3} instead of \texttt{r2}) or reorders commutative operands without altering the underlying operation. It also adjusts stack frame offsets or memory allocation sizes, provided that the modifications do not violate data dependencies or correctness constraints.

These variations suggest that the model is not merely memorizing instruction patterns but is instead learning high-level register-to-variable mappings and instruction equivalence classes. This flexibility enables generalization beyond the exact reference format and increases robustness to minor program transformations.

\begin{table}[h]
\small
\resizebox{\linewidth}{!}{\begin{tabular}{@{}p{0.8cm}p{0.8cm}p{6.4cm}@{}}
\toprule
\textbf{Prog ID} & \textbf{Edit Dist} & \textbf{Combined Patterns and Examples} \\
\midrule
P128 & 78 & \textbf{Multiple Optimization Patterns:} \\
& & Groud truth: \code{mul} \reg{r1}, \reg{r2}, \reg{r3} \\
& & Predicted: \\
& & \code{lsl} \reg{r1}, \reg{r2}, \imm{\#2}; \\
& & \code{add} \reg{r1}, \reg{r1}, \reg{r2} \\
\midrule
P113 & 74 & \textbf{Memory and Instruction Patterns:} \\
& & Ground truth: \\
& & \code{str} \reg{r1}, \mem{[fp, \#-12]} \\
& & \code{mov} \reg{r3}, \reg{r2} \\
& & \code{add} \reg{r3}, \reg{r3}, \imm{\#4} \\
& & Predicted: \\
& & \code{str} \reg{r1}, \mem{[fp, \#-8]} \\
& & \code{add} \reg{r2}, \reg{r2}, \imm{\#4} \\
\bottomrule
\end{tabular}}
\caption{Complex Variation Patterns with Multiple Differences}
\label{tab:complex_variations}
\end{table}

Furthermore, Table~\ref{tab:complex_variations} presents more substantial structural rewrites that nonetheless retain functional fidelity. These include compound transformations such as converting multiplications into equivalent shift-add sequences, or restructuring memory operations while preserving access order and scope. In one example, a multiplication instruction is replaced with a pair of shift and add instructions demonstrating the model’s awareness of performance-equivalent alternatives. In another case, memory writes and register arithmetic are reordered while maintaining the intended result, revealing the model’s competence in preserving state consistency across instruction sequences.

While these examples have higher edit distances, they exemplify a deeper form of equivalence: one grounded in operational semantics rather than surface-level syntax. The ability to produce such alternative forms underscores the potential of language models to reason compositionally about program structure and to synthesize diverse yet correct outputs for the same task.

In contrast, Table~\ref{tab:critical_errors} presents failure cases where minor syntactic deviations result in critical semantic errors. These include incorrect immediate values, register mismanagement, and mismatched memory offsets that compromise program correctness despite appearing superficially similar to the ground truth.

Together, Tables~\ref{tab:simple_variations},~\ref{tab:complex_variations}, and~\ref{tab:critical_errors} reveal that syntactic deviation does not necessarily imply failure. On the contrary, these examples support the argument that token-level metrics alone are insufficient to evaluate low-level transpilation tasks, and that functional correctness should take precedence in model assessment.

\label{sec:appendix}

\end{document}